# Generative Foundation Model for Structured and Unstructured Electronic Health Records

**Sonish Sivarajkumar, BS[1], Hang Zhang, MS[1], Yuelyu Ji, MS[1]; Maneesh Bilalpur, MS[1]; Xizhi Wu, MS[2]; Chenyu Li, MS[3]; Min Gu Kwak, PhD[2]; Shyam Visweswaran, MD, PhD[1,3,4], Yanshan Wang, PhD[1,2,3,4,5] †**

[1]*Intelligent Systems Program, University of Pittsburgh, Pittsburgh, PA*
[2]*Department of Health Information Management, University of Pittsburgh, Pittsburgh, PA*
[3]*Department of Biomedical Informatics, University of Pittsburgh, Pittsburgh, PA*
[4]*Clinical and Translational Science Institute, University of Pittsburgh, Pittsburgh, PA*
[5]*Hillman Cancer Center, University of Pittsburgh Medical Center, Pittsburgh, PA*
**† corresponding author: Yanshan Wang, yanshan.wang@pitt.edu**

## Abstract

Electronic health records (EHRs) are rich clinical data sources but complex repositories of patient data, spanning structured elements (demographics, vitals, lab results, codes), unstructured clinical notes and other modalities of data. Harnessing this heterogeneity is critical for improving patient outcomes - such as early detection of decompensation, timely intervention in chronic disease management, and reducing clinician documentation burden. Recent advances in large language models (LLMs) have enabled "foundation models" that can learn from multiple data modalities and support a spectrum of clinical tasks. However, most current approaches simply serialize numeric EHR data into text, which risks losing temporal and quantitative detail. We introduce Generative Deep Patient (GDP), a multimodal foundation model that natively encodes structured EHR time-series via a CNN – Transformer encoder and fuses it with unstructured EHRs through cross-modal attention into a LLaMA-based decoder. GDP is trained in two stages: (1) generative pretraining, where it learns to produce clinical narratives (e.g., brief hospital course) from raw patient timelines while also performing masked feature prediction (MFP) and next time-step prediction (NTP) to capture temporal dynamics; and (2) multi-task fine-tuning for clinically meaningful predictions (e.g., heart failure, type 2 diabetes, 30-day readmission).

In clinical prediction, GDP demonstrated superior performance on MIMIC-IV: heart failure AUROC = 0.923, type 2 diabetes AUROC = 0.817, and 30-day readmission AUROC = 0.627 . These results translate into earlier identification of high-risk patients, potentially enabling timely clinical intervention and resource allocation. For narrative generation, GDP achieved ROUGE-L = 0.135 and BERTScore-F1 = 0.545. In a blinded human evaluation, GDP-Instruct scored highest on faithfulness (4.4 / 5), fluency (4.6 / 5), and overall clinical utility (4.3 / 5), suggesting reduced hospital documentation workload without sacrificing accuracy.

Our results demonstrate that a single multimodal foundation model can both predict clinically actionable events (e.g., chronic disease onset, readmission risk) and generate high-quality clinical narratives, paving the way for more integrated AI models in everyday healthcare. Furthermore, GDP's flexible architecture can be extended to additional modalities (e.g., imaging, genomics) via their own encoders.

## Introduction

Electronic health records (EHRs), which compile longitudinal histories of patients from a variety of data modalities, have become a vast and transformative resource for clinical tasks Structured or codified data in EHRs ( e.g., demographics, diagnoses, procedures, vital signs, laboratory test results, and medications) and unstructured or free-text narratives together capture a comprehensive clinical picture. Leveraging this rich data for AI-driven insights poses significant challenges: the data are high-dimensional, longitudinal, irregularly sampled, heterogeneous, and often contain noise and missing values[1]. Traditional machine learning approaches struggle with the sheer volume and complexity of EHRs. A critical step is learning meaningful patient representations from these multimodal sources, which can enable accurate predictions of outcomes such as readmission, or new diagnoses[2].

The emergence of foundational models and generative models have spurred their application to healthcare. Deep models are neural networks with multiple hidden layers that learn hierarchical representations of data. The term "deep" refers to the fact that these models capture complex patterns by learning features at different levels of abstraction. Examples include deep convolutional networks for image recognition. Foundation models are large-scale models trained on broad, diverse data that can be adapted to many tasks[3]. They serve as the foundation for multiple applications through techniques such as fine-tuning or prompting. Examples include large language models(LLMs) like GPT. Generative models are designed to learn the underlying data distribution and generate new data that resemble the training data. Examples include variational autoencoders that learn compressed representations. Most modern generative and foundation models are deep models. Many foundation models are generative (like GPT), but not all (BERT is primarily discriminative). A model can be all three (GPT-4 is deep, generative foundation model).

In the context of EHRs, researchers have explored two main directions: (1) Clinical language models are trained on clinical free text narratives to understand and generate clinically relevant language (e.g., CLaMs) [4, 5], and (2) Foundation models learn patient representations from structured EHR timelines and can be adapted (via fine-tuning or prompting) to perform specific clinical tasks (e.g., FEMRs) [6, 7]. While CLaMs treat input as tokens, structured EHR data is inherently numerical, so transforming structured EHR into tokens and then encoding them back into an LLM decoder might not be the most effective approach. Moreover, converting numerical and time-series EHR elements into text tokens for LLMs can degrade numeric precision and temporal context - factors that are often critical for patient safety and risk stratification. FEMRs, on the other hand, produce dense patient embeddings from a patient's entire record, which can feed downstream predictive models. Studies have shown that such EHR-specific foundation models can improve predictive accuracy and require fewer labeled examples than traditional methods. For instance, self-supervised models trained on large EHR datasets have achieved state-of-the-art results on clinical risk prediction while maintaining robustness to shifts in data distribution[8]. At the same time, general-purpose LLMs like GPT have demonstrated impressive language understanding and medical knowledge, but often underperform domain-specific models, especially on structured prediction tasks. Specialized biomedical language models, such as BioBERT[9] and Gatortron[10], have been developed to better capture medical language compared to general language models. However, they lack the generative capabilities of LLMs and are not optimized for question answering tasks.

One recently proposed approach to bridge structured EHR data with LLMs is to serialize patient records into textual form that LLMs can interpret[11, 12]. In this paradigm, as in CLaMs, all structured elements (e.g. codes, time-stamped events) are converted into a formatted text (such as a lengthy Markdown document) with human-readable descriptions. An LLM (or an LLM-based embedding model) processes this text to yield a patient embedding. This strategy leverages LLMs' broad knowledge and avoids needing a proprietary EHR corpus for initial training. However, this serialized-text approach has several limitations. First, it converts numerical variables into text, and the LLM encodes it into numerical vectors, during which the data loses various contexts and details[13]. Second, it relies on a hand-crafted schema for representing structured data as text, which may introduce biases or omit subtle information[14]. Model performance can be sensitive to the exact formatting and phrasing of the serialized record, raising reproducibility concerns. Third, processing long textual records pushes against LLM context length limits - important historical data may be truncated, especially for patients with lengthy records[15]. Perhaps most critically, when the LLM's output is used only as features for a separate classifier (for example, a logistic regression on the LLM embedding), we lose the LLM's native ability to perform reasoning or generation in the final step of the task. In other words, the pipeline forfeits the zero-shot and conversational capabilities of the LLM at the point of making clinical predictions. These observations motivate a unified multimodal framework that treats structured EHR and clinical text as complementary inputs - each encoded in its native form - rather than flattening all signals into text.

An alternative paradigm, supported by extensive prior work, is multimodal learning: designing models that explicitly integrate each data modality in its native form[16]. Instead of flattening everything into text, a multimodal architecture can use appropriate subnetworks for structured numerical data, time-series signals, and text, and then fuse these representations. By preserving each modality's structure, such models can capture patterns that might be lost in text translation. Multimodal fusion approaches have shown improved performance on clinical predictions by combining complementary information sources. For example, earlier deep learning systems combined coded data and clinical notes via separate encoders (e.g. recurrent networks or convolutional neural networks for sequences of codes and transformers[17] for text) and achieved higher accuracy than single-modality models. This suggests that a holistic model, ingesting the full spectrum of EHR data, could learn more robust patient representations and yield better predictions than models that treat the record as only text or only codes.

Inspired by these, we propose Generative Deep Patient (GDP), a new multimodal foundation model framework for both structured and unstructured EHRs. GDP is designed to synthesize the strengths of multimodal integration and LLMs. The GDP architecture (Figure 1) employs dedicated encoders for structured EHR time-series data and for unstructured text, and then fuses their outputs into a unified latent representation using an LLM backbone with cross-modal attention[18]. This design allows the model to absorb the rich temporal patterns of structured data as well as the semantic context of clinical text. Uniquely, we explore two variants of the LLM backbone: one initialized with an instruction-tuned model (GDP-Instruct) and one with a standard pre-trained model (GDP-Base)[19]. Both GDP variants undergo a two-stage training regime. First, a generative pretraining stage teaches the model to generate clinical narrative conditioned on the patient's structured data, while simultaneously employing auxiliary self-supervised objectives to learn the temporal dynamics of the EHR. Second, a multi-task fine-tuning stage adapts the model to specific predictive tasks (e.g., readmission, diagnosis prediction, and so on) using labeled data, while leveraging the learned representations. Unlike the purely text-serialized LLM approach, our framework aims to learn a unified multimodal patient representation within the model, retaining the foundation model's expressive power all the way through to the final task output.

**Figure 1.**

A generalist multimodal architecture for clinical foundational models. In this paper, we focus on a foundation model combining the first two modalities, i.e., structured EHRs and unstructured free-text EHRs. Discriminative tasks - directly predict specific outcomes or labels from patient data (e.g., diagnosing HF or estimating readmission risk). Generative tasks - produce new clinical data, specifically text in this study, (e.g., discharge summaries or progress notes).

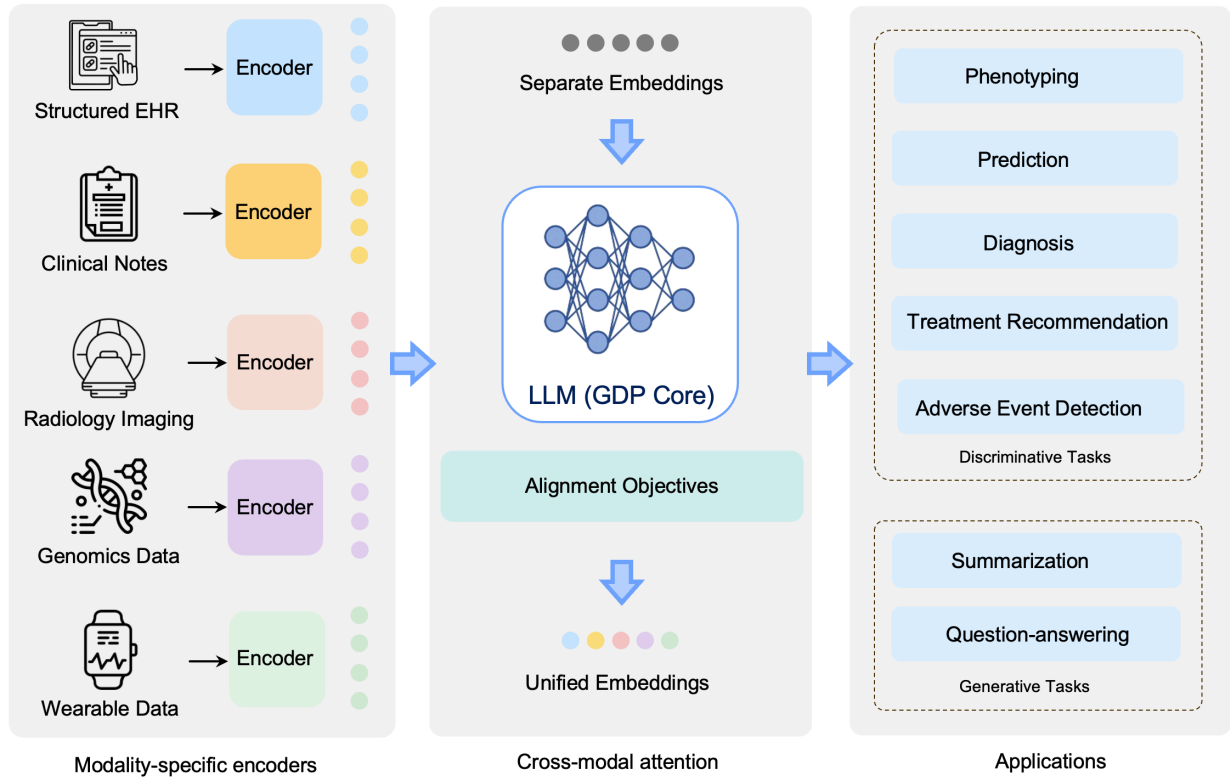

We hypothesize that explicit multimodal processing, using separate data encoders, with a fused LLM backbone will produce more robust and generalizable patient representations than methods that serialize all data into text for an LLM. In particular, the inclusion of auxiliary temporal modeling tasks - namely masked feature prediction and next time-step prediction - during pretraining should imbue GDP with a deeper understanding of clinical time series, conferring an advantage on sequential prediction problems. The key objective of this work is to develop a generalist foundational model framework – GDP - capable of capturing the representations from multiple modalities of data. Additionally, the work aims to evaluate GDP's performance across multiple modalities and task types, and to assess whether an instruction-tuned multimodal model can serve as a generalist AI system for EHR that both predicts clinical outcomes and explains patient history in natural language.

# Results

GDP is designed to perform well in both discriminative and generative tasks. A discriminative task is one where the model learns to predict a specific outcome or label from patient data. For example, predicting HF or assessing the readmission of a patient after hospital discharge. Discriminative models directly approximate the conditional probability P(label|data) and are evaluated by classification metrics (e.g., AUROC, AUPRC, F1). In contrast, a generative task is one where the model learns to produce new, realistic clinical text based on the same patient data,such as generating a "Brief Hospital Course" summary. Generative models approximate the joint or conditional probability of the text given the data, P(text|data), and are evaluated by language-generation metrics (e.g., ROUGE, BLEU, BERTScore) or human assessment of fluency and faithfulness.

We evaluated GDP in three phases:

1. Clinical prediction tasks (discriminative): We selected three key outcomes - heart failure (HF), type 2 diabetes mellitus (T2DM), and 30-day readmission - because they represent high-impact conditions where early identification can change management (e.g., adjusting diuretics for HF, tighter glycemic control for T2DM, care-transition planning to reduce readmissions).

2. Clinical narrative generation (generative): We evaluated how accurately the model can produce the "Brief Hospital Course" section of discharge summaries, since concise, accurate hand-offs are critical for continuity of care.

3. Ablation and subjective evaluation: We performed ablation to isolate the contributions of auxiliary objectives (masked feature prediction using label "[MFP]" and next time-step prediction using label "[NTP]") and "instruction tuning," as well as a blinded human rating to assess faithfulness and utility(Supplementary Material Section 4 and 5).

Below, we first present results on the discriminative tasks (Table 1, Figure 2), then on narrative generation (Table 2, Table 3, Figure 3), followed by ablation analyses and qualitative chat demonstrations

## Predictive performance on clinical tasks

We evaluated the fine-tuned GDP models on three key prediction tasks derived from the hospital EHR data: identification of two prevalent chronic conditions (HF and T2DM) based on data available early in the admission., and 30-day hospital readmission.  We compared GDP against three existing approaches: (a) CEHR-BERT, a Transformer-based EHR sequential model[20], (b) FPM (SDAE)[8], a stacked denoising autoencoder on EHR data with logistic regression based on the deep patient model[21], and (c) CLMBR, a recent self-supervised EHR foundation model (141M parameters) that autoregressively predicts the next code in a patient timeline[6]. We selected these approaches to span different modeling strategies - Transformer-only, autoencoder-plus-classifier, and autoregressive EHR models, that GDP's multimodal fusion and pretraining approach could be fairly compared against diverse, high-performing alternatives. Figure 2 illustrates the AUROC and AUPRC curves for the models on each task.

> *Key terms:*
> *• Transformer - a deep learning architecture that uses self-attention to model relationships between events or tokens in a sequence.*
> *• Autoencoder - a neural network that compresses data into a lower-dimensional representation (encoder) and reconstructs it (decoder), often used for feature learning.*
> *• Classifier - a model component that assigns an input to a discrete label (e.g., presence of heart failure vs. not).*
> *• Autoregression - a modeling approach where the next element in a sequence is predicted based on its past values.*

**Figure 2.**
Discriminative performance across clinical prediction tasks.
(a) AUROC and (b) AUPRC for T2DM, HF, and 30-day readmission, comparing GDP and three baseline models.

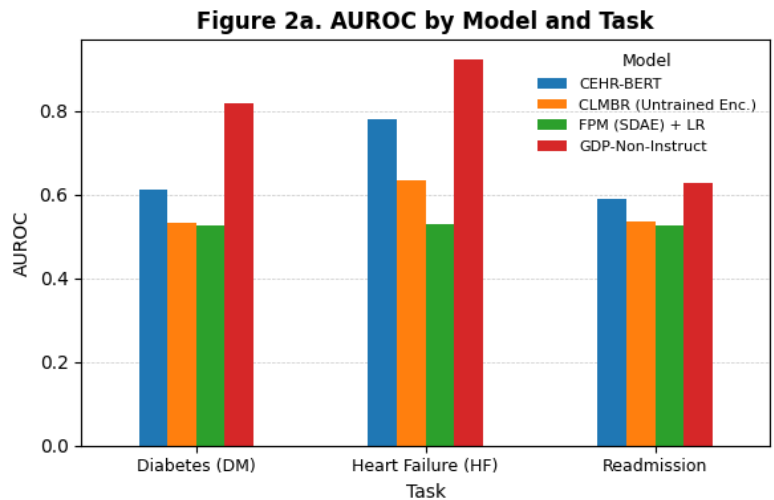

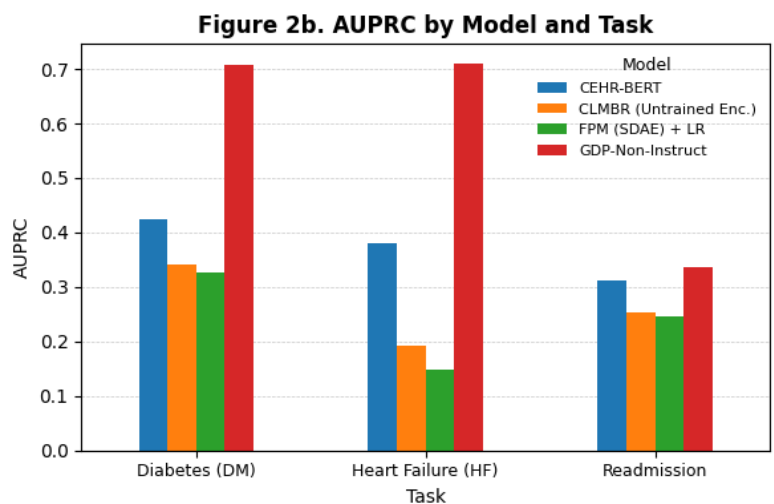

Figure 2b. AUPRC by Model and Task

**Table 1.**

Discriminative performance metrics on EHR prediction tasks. F1-score, precision, recall, and accuracy for GDP variants and baseline models (CEHR-BERT, FPM+LR, CLMBR) on HF (ICD-9 428.x), T2DM (ICD-9 250.x), and 30-day readmission.

| Model | Task | F1-Score | Precision | Recall | Accuracy |
|---|---|---|---|---|---|
| CEHR-BERT | ICD428 (HF) | 0.2554 | 0.55 | 0.17 | 0.87 |
| | ICD250 (T2DM) | 0.1323 | 0.62 | 0.07 | 0.70 |
| | Readmission | 0.0312 | 0.61 | 0.03 | 0.77 |
| FPM (SDAE) + LR | ICD428 (HF) | 0.2317 | 0.13 | 0.79 | 0.13 |
| | ICD250 (T2DM) | 0.4657 | 0.30 | 0.70 | 0.30 |
| | Readmission | 0.430 | 0.35 | 0.55 | 0.77 |
| CLMBR | ICD428 (HF) | 0.2854 | 0.18 | 0.66 | 0.57 |
| | ICD250 (T2DM) | 0.3956 | 0.33 | 0.50 | 0.53 |
| | Readmission | 0.3336 | 0.25 | 0.52 | 0.52 |

|  | ICD428 (HF) | 0.6020 | 0.47 | 0.84 | 0.85 |
| --- | --- | --- | --- | --- | --- |
| GDP-Non-Instruct | ICD250 (T2DM) | 0.6218 | 0.55 | 0.72 | 0.74 |
|  | Readmission | 0.3967 | 0.30 | 0.57 | 0.60 |

Overall, the GDP models achieved the highest discriminative performance. For HF prediction, GDP-Non-Instruct obtained an AUROC of 0.923 (95% CI ~0.91–0.93) and AUPRC of 0.71, substantially outperforming the best baseline (CEHR-BERT, AUROC 0.779, AUPRC 0.38) and the CLMBR model (AUROC 0.64, AUPRC 0.19). Its F1-score was 0.60, far higher than the baseline CEHR-BERT's 0.25 (the latter struggled with very low recall of 0.17 for HF). T2DM prediction showed a similar pattern: GDP-Non-Instruct reached AUROC 0.817 and AUPRC 0.707, versus 0.610 and 0.424 for CEHR-BERT. The simpler baselines (SDAE and CLMBR) performed near chance on HF and T2DM (AUROCs ~0.53–0.63). On the more challenging 30-day readmission task, all models had lower scores, reflecting the inherent difficulty of predicting readmissions. GDP-Non-Instruct again led with AUROC 0.627 and AUPRC 0.335, compared to 0.589 and 0.311 for the best baseline (CEHR-BERT). Notably, CEHR-BERT's F1 on readmission was only 0.03 (it almost never predicted the positive class, yielding recall 1.6%), whereas GDP-Non-Instruct achieved F1 ≈0.40 by balancing precision and recall (~0.30 and 0.57, respectively). The differences between GDP-Non-Instruct and the best baseline (CEHR-BERT) were statistically significant in the HF and T2DM tasks. In readmission, the improvement was modest. These results demonstrate that GDP's multimodal fusion and pretraining strategy conferred a sizable advantage over prior EHR-specific models in two of three evaluated clinical predictions. We attribute this to GDP's ability to capture long-term temporal patterns (through its CNN–Transformer encoder and NTP objective) and to integrate information across notes and coded data. The instruction fine-tuning of the LLM, however, was not necessary (and perhaps slightly disadvantageous) for maximizing structured-task performance in this setting.

### Generative performance on clinical text

We next evaluated each model's ability to generate specific sections of a discharge summary from structured EHR inputs. In this simulated setup, models receive a patient's admission events and context and must produce the concise narrative clinicians write at discharge. We fine-tuned both GDP-Instruct and GDP-Non-Instruct on paired structured data and ground-truth summary text, and compared them against three existing models: (1) a 3B-parameter LLaMA-derived model in non-instruct ("Meta Llama Non-Instruct") and instruct-tuned ("Meta Llama Instruct") forms[22], and (2) Med42-a 8B clinical LLM based on Llama-2, instruction-tuned on medical[23]. These three baselines were chosen to have a comprehensive comparison on non-instruct(base), instruct and a finetuned versions of LLMs.

We report standard automated language-generation metrics-ROUGE-1, ROUGE-2, and ROUGE-L F-scores (n-gram overlap measures), BLEU-4 (precision-oriented n-gram overlap), and BERTScore-F1 (semantic similarity)-for all model outputs (Table 2). Higher values indicate closer agreement with the reference clinician-written summaries.

**Table 2.**
Automated generation metrics for discharge summary. ROUGE-1, ROUGE-2, ROUGE-L F1-scores, BLEU-4, and BERTScore-F1 for baseline LLaMA models (instruct/non-instruct), Med42, and both GDP variants.

| Model | ROUGE-1 (F) | ROUGE-2 (F) | ROUGE-L (F) | BLEU-4 | BERTScore-F1 |
|---|---|---|---|---|---|
| Meta Llama-Non-Instruct | 0.11 | 0.01 | 0.08 | 0.01 | 0.44 |
| Meta Llama Instruct | 0.12 | 0.01 | 0.08 | 0.01 | 0.46 |
| Med42 | 0.17 | 0.03 | 0.11 | 0.01 | 0.5 |
| GDP-Non-Instruct | 0.2 | 0.04 | 0.13 | 0.01 | 0.54 |
| GDP-Instruct | 0.21 | 0.05 | 0.14 | 0.02 | 0.55 |

GDP-Instruct demonstrated superior performance on every metric-ROUGE-1 of 0.215, ROUGE-2 of 0.048, and ROUGE-L of 0.135-demonstrating superior n-gram overlap with the reference summaries. Its BLEU-4 score (0.017) further indicates more accurate reproduction of multi-word sequences, and its BERTScore-F1 (0.545) confirms best semantic fidelity among the models. GDP-Non-Instruct ranked second, outperforming the much larger Med42 model (ROUGE-L = 0.105, BLEU-4 = 0.009, BERTScore = 0.500) despite having only 3 billion parameters. The instruct-tuned Meta Llama baseline showed only modest gains over its non-instruct version (ROUGE-L = 0.075 vs. 0.075, BERTScore = 0.460 vs. 0.435), underscoring the benefit of both instruction tuning and our EHR-focused pretraining.

A ROUGE-L F1 of 0.135 indicates that the model captures approximately 13.5 % of the longest common subsequence between its output and the reference "Brief Hospital Course," which may seem low by general NLP standards—but in clinical practice, even capturing key events (e.g., "IV antibiotics started," "elevated troponin noted," "discharged on beta-blocker") can substantially reduce clinician documentation time. These quantitative results highlight that targeted pretraining on clinical data, combined with instruction optimization, yields high-quality, clinically faithful summaries without requiring enormous model scale.

**Table 3.**

Mean human evaluation scores (1 – 5 Likert) for generated discharge summaries. Ratings for faithfulness, completeness, conciseness, fluency, and clinical utility averaged over two expert annotators; inter-annotator agreement κ = 0.68

| Model | Faithfulness | Completeness | Conciseness | Fluency | Utility |
|---|---|---|---|---|---|
| GDP-Instruct | 4.4 | 4.2 | 4.0 | 4.6 | 4.3 |
| GDP-Non-Instruct | 4.1 | 3.3 | 3.6 | 4.1 | 3.5 |
| Med42 | 4.0 | 3.6 | 3.8 | 4.4 | 3.8 |
| Meta Llama Non-Instruct | 3.6 | 3.1 | 3.1 | 4.0 | 2.8 |
| Meta Llama Instruct | 4.0 | 3.5 | 3.8 | 4.3 | 3.7 |

In addition to automated metrics, we conducted a blind human evaluation of the generated summaries to assess their clinical usefulness and correctness, adhering to QUEST guidelines[24] (Table 3). Two health informatics specialists reviewed a randomized sample of 200 summary outputs per model, rating each on five criteria, as defined by QUEST human evaluation framework [24]: Faithfulness, or how accurately the summary reflects the patient's clinical facts; Completeness, meaning coverage of all key clinical events; Conciseness, which captures essential information without unnecessary detail; Fluency, referring to grammatical smoothness and readability; and Overall Clinical Utility, indicating how useful the summary would be for real-world decision-making. Ratings used a 1–5 Likert scale (5 = best), and evaluators were blinded to the model source. Inter-annotator agreement was substantial (overall Cohen's κ = 0.68), with κ by criterion as follows: Faithfulness 0.62, Completeness 0.64, Conciseness 0.66, Fluency 0.75, and Utility 0.60. The scores suggest that GDP can generate high quality data while maintaining predictive accuracy.

**Figure 3.**

Human evaluation radar chart. Mean Likert ratings (1 – 5 scale) for faithfulness, completeness, conciseness, fluency, and clinical utility of discharge-summary generation outputs, averaged over two expert evaluators (n = 200 summaries per model).

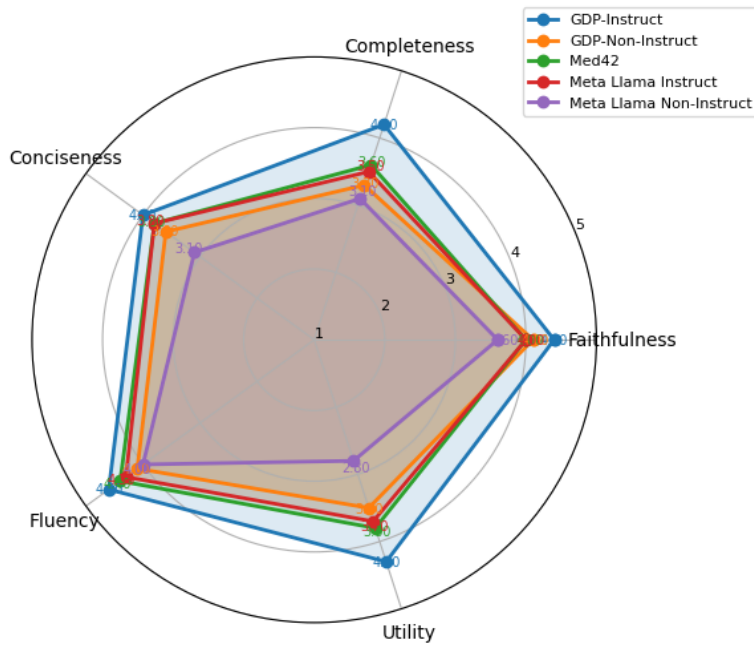

GDP-Instruct achieved the highest mean rating on four of five criteria-most notably Fluency (4.6) and Faithfulness (4.4)-and matched the top Conciseness score (4.0). It outperformed the next-best model (Meta Llama Instruct) by 0.5 points on Completeness and Utility. The evaluators noted that GDP-Instruct outputs "generally stick to documented facts without hallucinating," whereas other models occasionally omitted or misstated key events. GDP-Non-Instruct performed reasonably but lagged behind GDP-Instruct across all dimensions, particularly in Utility (3.5), reflecting less coherent and actionable narratives without instruction tuning. Figure 3 presents these results in a radar chart, illustrating GDP-Instruct's uniformly stronger performance across clinical evaluation axes. These findings confirm that instruction tuning on EHR-derived tasks substantially enhances both the factual reliability and practical value of generated discharge summaries.

## Discussion

In this work, we introduced Generative Deep Patient (GDP), a multimodal foundation model that bridges the gap between predictive analytics and generative understanding in EHRs. Current foundation models in healthcare primarily focus on unstructured data, such as clinical notes and medical literature[25, 26]. However, they often overlook the vast amount of structured EHR data available. This limitation creates a significant research gap in leveraging the full potential of healthcare data for improved patient outcomes and clinical decision-making. GDP provides a more comprehensive approach to healthcare analytics and predictive modeling. This innovative approach enables healthcare professionals and researchers to gain deeper insights into patient health patterns, treatment efficacy, and population-level trends. Furthermore, GDP's ability to process structured EHR data alongside unstructured information enhances the model's accuracy and applicability in real-world clinical settings, potentially leading to more personalized and effective patient care strategies.

The results demonstrate that a single model can indeed perform well on both structured outcome prediction and generative tasks. In GDP-Instruct, instruct LLM backbone translated to superior performance in the summary generation and a smoother conversational ability. The model could leverage its instruction-following training to organize information from the EHR into a fluent summary or answer user queries effectively. On the other hand, GDP-Non-Instruct's backbone remained closer to a standard language model that optimizes next-token prediction on general text. After fine-tuning on our supervised tasks, it appeared to retain a sharper focus on the discriminative signals in the data, yielding higher AUROC/AUPRC. One hypothesis is that the instruction-tuned model had been predisposed to prioritize natural language coherence and high-level reasoning, which might introduce a form of regularization (or even a slight distraction) when it is repurposed for strict classification objectives. The non-instruct model, being more "raw", might more readily memorize and exploit subtle statistical patterns in the structured data during fine-tuning, hence its edge in predictive performance. This echoes observations in other multimodal domains where instruction-tuned vision-language models trade some accuracy for better alignment and usability[27].

Importantly, both GDP variants significantly outperformed existing baselines in most metrics, indicating that our multimodal fusion strategy and pretraining regimen are fundamentally sound and beneficial. The incorporation of temporal modeling tasks (NTP, MFP) during pretraining is a key strength[28]. These tasks forced the model to learn the underlying structure of the patient timeline (e.g. that certain events follow others in time, that certain labs correlate with specific diagnoses, etc.), rather than treating the structured data as an uncorrelated bag of features. This likely contributed to the robust performance of GDP-Non-Instruct in predictions , unlike prior EHR Transformers (CEHR-BERT) that only rely on sequence position embeddings, GDP's encoders explicitly learned to predict unseen or next events, imbuing a sense of chronological order and progression. Even for GDP-Instruct, which ultimately shined in text generation, those same pretraining tasks improved the factual grounding of its outputs. We see evidence of this in the human evaluation: GDP-Instruct had remarkably high faithfulness scores, suggesting it seldom hallucinated or omitted key facts. We attribute this to the model's cross-attention mechanism which ties generation to the patient's data, and to the auxiliary objectives that ensured the structured data representation is rich and descriptive of the patient. In contrast, a large generic LLM (even one tuned on medical text) might produce a fluent summary but could wander or embellish if not tightly anchored; GDP-Instruct, having been trained to generate from data, was less prone to such errors. This highlights an attractive aspect of our approach: by retaining the "foundation model" training within the multimodal domain, we combine the accuracy of specialized models with the flexibility of LLMs. The model effectively learns an internal medical representation that it can use for both making predictions and generating explanations-a step toward AI that is both right and interpretable.

Despite these strengths, our study has several limitations that suggest avenues for future work. First, the context length of the LLM (and the practical need to limit input size to 128k) constrained how much of a patient's history we could use. In fine-tuning, we restricted structured sequences to 50 time steps and focused on a single key clinical note (e.g. discharge summary) per admission. Patients with longer or more complex histories may have relevant information beyond these limits. This is a general challenge as EHRs can span hundreds of events and dozens of notes - future models might incorporate longer context handling (through hierarchical encoding[29], or retrieval mechanisms[30]) to ensure no pertinent data are dropped. Second, our evaluations were on a single-center dataset (MIMIC-IV) of ICU and hospital inpatients. Relatedly, we only tackled three prediction tasks (readmission, two diagnoses) and one type of summarization. These were chosen as representative tasks, but they do not cover all possible clinical questions. The model's utility on other targets (e.g. length of stay regression, adverse event prediction, procedure recommendation, etc.) should be explored. Third, model size and efficiency are considerations. Our LLM backbone was 3.2 billion parameters – relatively small by modern standards – primarily due to computational constraints in training a multimodal model end-to-end. It is possible that larger LLMs (e.g. 8B, 13B, or beyond) would yield even better performance, especially on generative tasks, as suggested by the baseline scaling trends (the 8B model was competitive, though it lacked our EHR-specific pretraining). However, training and deploying such large models in healthcare settings is challenging. Future research might investigate knowledge distillation or adapter-based tuning to compress the GDP model or deploy it in parts (for instance, a lightweight encoder at the bedside that feeds into a cloud-based LLM). Additionally, the current architecture uses a full cross-attention integration of structured data at each layer of the LLM, which is computationally expensive. More efficient fusion mechanisms (such as intermittent cross-attention or gating, or using an encoder–decoder design where the decoder attends to a fixed embedding sequence) could reduce latency and memory usage, which are important for practical use.

Another limitation is that our multimodal scope was restricted to structured EHR data and text. EHRs often contain other data types like medical images (radiology scans), waveforms (ECG, EEG), and possibly genomics[31, 32]. GDP's architecture could, in theory, be extended with additional encoders for these modalities (and some recent works do explore multimodal foundation models including imaging), but we did not incorporate or test those. Therefore, our assertion of being "generalist" is somewhat constrained. An ideal clinical foundation model would seamlessly incorporate imaging and other modalities. However, structured EHR is frequently neglected and not extensively studied, which is what makes GDP both adaptable and innovative. Incorporating imaging data (for example, having the model also process chest X-rays when generating the summary or predicting outcomes) is a promising direction for making the model even more comprehensive. Similarly, future models might ingest genomic profiles or other patient-specific data to enrich predictions for precision medicine.

Looking ahead, several future directions emerge. One is to explore the zero-shot and few-shot learning potential of GDP. As an instruction-tuned LLM fused with clinical data, GDP-Instruct might be able to handle new tasks via prompting, without additional fine-tuning, as shown in many existing studies[33, 34]. For example, given a patient record, could we prompt it with an instruction like, "List this patient's active problems and suggested management plan," and get a sensible result? This would move toward more flexible AI assistants that don't require a new supervised model for every task. Preliminary chat experiments are encouraging in this regard, but systematic evaluation on benchmark clinical Q&A datasets or few-shot prediction tasks (like those in the recent EHRSHOT challenge would be informative. Another direction is to incorporate reinforcement learning from human feedback (RLHF) to further refine the model's generative behavior[35]. Our human evaluation gave us insight into what clinicians prefer; using that as a signal to fine-tune GDP-Instruct (similar to how ChatGPT is refined) could improve its utility and safety in an interactive setting. Additionally, exploring modular or lightweight encoder updates is worthwhile. For instance, if a hospital has an existing GDP model, how can it be efficiently adapted to local data? Techniques like Low-Rank Adaptation (LoRA) or adapter modules inserted into the GDP architecture might allow fine-tuning on a new dataset without retraining the whole model[36]. This would be particularly useful for sharing a base model across institutions, letting each site customize it with a small amount of local data – a scenario that early studies on foundation models for EHR have started to examine. Finally, it will be important to test GDP in a prospective simulation or clinical trial context: e.g., have the model generate summaries or risk predictions on live hospital data and measure outcomes such as physician satisfaction, documentation time saved, or diagnostic accuracy when assisted by the model. These real-world evaluations will ultimately determine if models like GDP can transition from research proof-of-concept to trustworthy tools in everyday healthcare.

# Methods

**Figure 4.**

Overview of data processing and experimental pipeline.

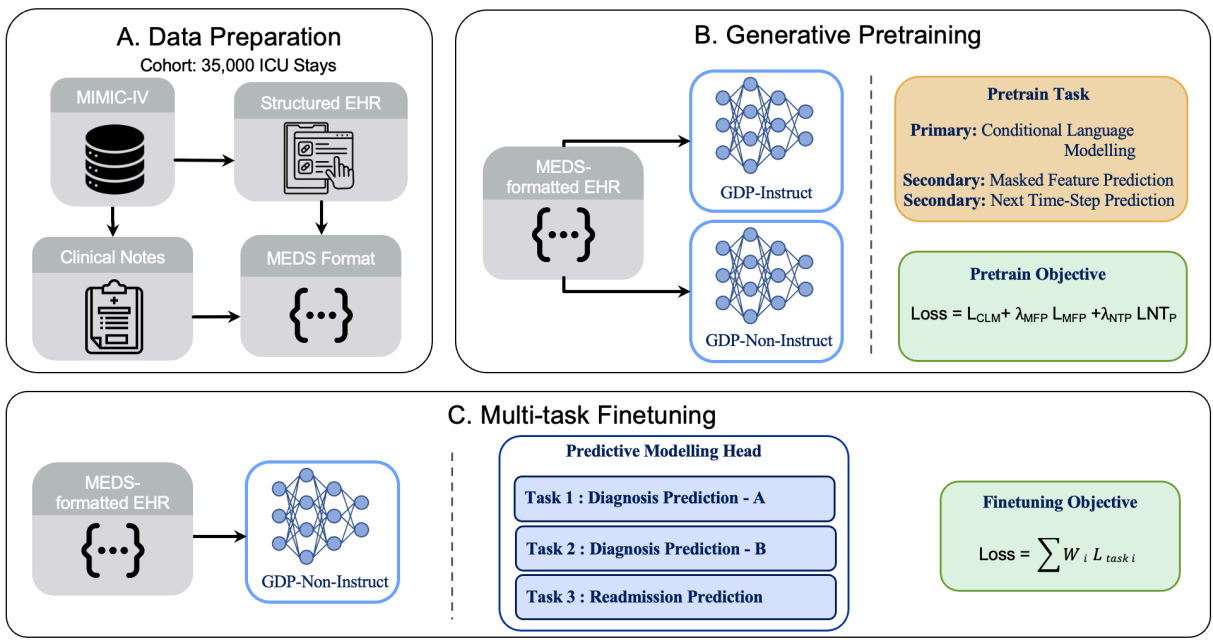

## Data source and cohort

We developed and evaluated our model using the Medical Information Mart for Intensive Care IV (MIMIC-IV) dataset that is available through PhysioNet with credentialed access. MIMIC-IV comprises de-identified health records for patients admitted to the ICU at a large academic medical center. For this study, we used a subset of MIMIC-IV, containing data from ~35,000 hospital admissions between 2008 and 2019[37]. Each patient encounter was represented as a combination of structured events and text. We prepared the data in a Medical Event Data Standard(MEDS) format (inspired by prior MIMIC-Extract pipelines), where each line is a JSON record containing all relevant information for one patient admission[38]. In structured data, we included: demographics (age, sex), diagnoses and procedures (ICD-9, CPT-4), laboratory test results (LOINC), medication administrations (RxNorm), and measurements such as vital signs, input/output and ventilator settings. All of these data are time-stamped. Each event entry consisted of a timestamp and a feature (or code) with an associated value (for labs/vitals) or category. Each unit of analysis corresponded to a single hospital admission. Patients could contribute more than one admission, but splits into training, validation, and test were made at the patient level so that no admissions from the same patient appeared across different sets. These events were sorted chronologically to form the patient's timeline. In text data, we included the content of clinical notes – specifically, we leveraged discharge summaries as they provide a synopsis of the entire hospital stay.

## Discriminative and generative tasks

For training the generative model, we parsed discharge summaries to extract the Brief Hospital Course (BHC) section (a narrative paragraph summarizing the hospitalization). The BHC section was used as the target text for generation. For multi-task prediction, we defined three binary outcome labels based on the EHR data: 30-day readmission (unplanned readmission to hospital within 30 days of discharge), HF diagnosis (whether the patient had a diagnosis code for HF, ICD-9 428.x, during the admission), and T2DM (ICD-9 250.x for type 2 diabetes). A positive label for HF or T2DM indicates that condition was present (per coding or problem list) during the hospitalization. For prediction tasks, we used data available at the time of admission (early hospital events) to predict HF, DM, or 30-day readmission risk. For narrative generation, the model used the entire admission timeline to generate the BHC at discharge. The remaining tasks had a mix of positive and negative cases (prevalence: HF ~11%, T2DM ~14%, readmission ~15%). We split the dataset into training, validation, and holdout test sets at the patient level (so no patient's records appear in more than one set). The test set was about 15% of the data (thousands of admissions for each task). All structured features were normalized or encoded before input: continuous values (lab results, etc.) were z-scored or bucketed, and categorical codes were mapped to integer indices. Table X in the supplementary material lists the data types and encoding used.

## Data preprocessing

All data were processed through a standardized pipeline to prepare model inputs. Structured event sequences: Each patient's timeline of events was transformed into a fixed-length sequence of vectors for modeling. We first defined a vocabulary of the most frequent coded events (diagnoses, procedures, medications, lab types, etc.) and allocated an embedding vector for each code. For each timestamp in a patient's record (e.g. each hour or each day with recorded data), we aggregated the events occurring at that time into a single vector. Specifically, if multiple codes occurred at the same time, their embeddings were averaged; associated numeric values (like lab results) were appended as scaled features. We also appended time features such as the time since admission for each event. If a sequence had more events than the maximum length (we set max length = 50 for fine-tuning, based on distribution of stay lengths), we truncated it to the most recent 50 events; if shorter, we padded it with dummy vectors. This resulted in a [$T \times d$] matrix per patient (with

$T$ = 50, $d$ = embedding dimension, e.g. 128), representing the structured data timeline. We applied minor data cleaning, for example filtering out rarely-used codes and forward-filling missing values for vitals within short windows. Text data: The discharge summary (or specifically the BHC section) was tokenized using a subword tokenizer (the same tokenizer as the LLM backbone). We limited the length to a set number of tokens (e.g. 256 tokens) to fit the model's context window. In generative pretraining, the structured data sequence served as context and the full BHC text was the target to generate. For fine-tuning tasks, we also processed notes: for each admission, we took the text of the discharge summary (minus the BHC) as additional input to the model's text encoder (described below). All text was lowercased and de-identified (MIMIC is already de-identified; we further removed any surrogate identifiers). No other text normalization was done; the clinical language was kept in its original form to preserve meaning.

## Model architecture

The GDP model is a multimodal encoder–decoder architecture built around a Transformer-based LLM. Figure 1 depicts its components. There are three main subsystems: (1) the Structured Data Encoder, (2) the Unstructured Text Encoder, and (3) the LLM backbone with multimodal fusion. Additionally, the model has task-specific output heads for generation and classification.

**Figure 5.**

Detailed GDP model components.

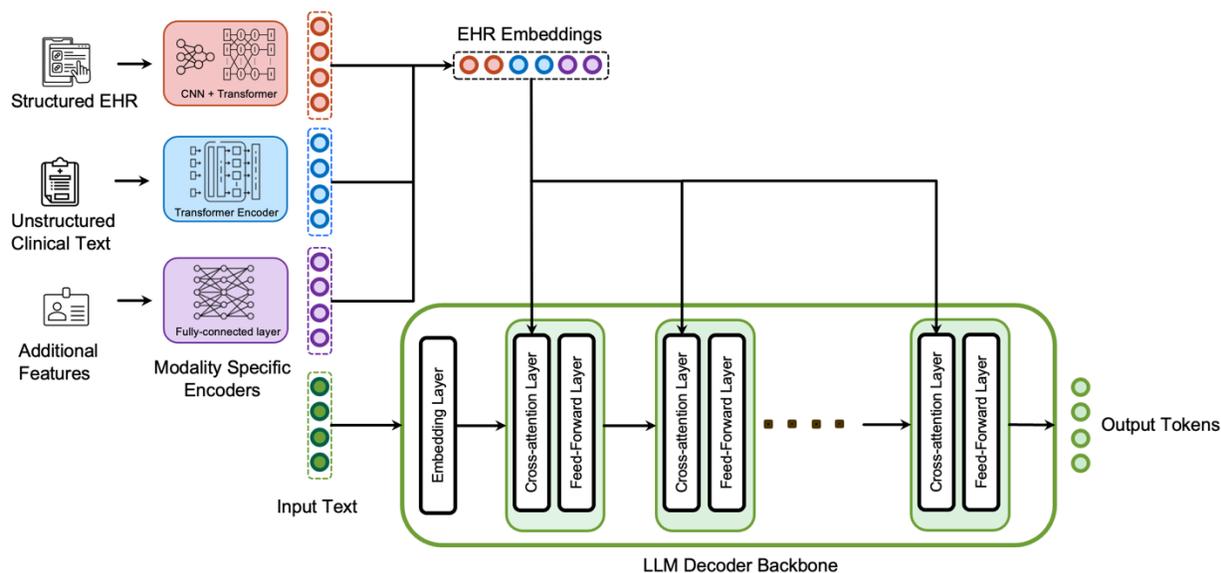

**Structured Data Encoder:** This module processes the sequence of structured EHR event vectors. It begins with several 1-D convolutional layers that scan over the time dimension of the event sequence. These CNN layers capture local temporal patterns and smooth out high-frequency noise in the sequences (for example, detecting a short trend in vital signs)[39]. Next, the encoder includes Transformer encoder layers (self-attention blocks) that operate over the sequence length $T$. The Transformer layers enable learning of long-range dependencies across the entire timeline, for instance, attention can connect an event early in the hospitalization with one much later, if relevant. We used positional encodings to preserve the order of events in these attention layers. The structured encoder produces a sequence of hidden vectors of length $T$. We projected the final outputs to a hidden dimension of 256 (so each time step is represented by a 256-d embedding). If present, a separate small Feature Encoder processes any auxiliary features (like static demographics or time-gap features) via a simple feed-forward network, and outputs a 256-d embedding that can be added to or concatenated with the main sequence representation. The design choice of a hybrid CNN+Transformer for structured data balances efficiency and expressiveness: CNN layers rapidly reduce noise and extract short-term trends, while Transformers capture global temporal context. This is advantageous over relying solely on recurrent networks, which can be slower and may struggle with long sequences. More details about the architecture are provided in the Supplementary material.

**Unstructured Text Encoder:** For clinical text inputs (e.g. discharge notes), we employed a pretrained clinical BERT encoder. Specifically, we used the BioClinicalBERT model (a BERT-base model further pre-trained on biomedical and clinical text) as our text encoder[4]. This encoder takes tokenized text (up to ~512 tokens) and produces a contextual embedding. We take the [CLS] token embedding (a 768-d vector) from the BERT encoder as a summary of the note's content. A linear projection then maps this to the 256-d hidden space. During model training, this text encoder can be fine-tuned or kept frozen depending on the stage – in generative pretraining, we did update it when generating text conditioned on text+structured data; in fine-tuning, we typically froze BERT initially and then allowed fine-tuning in later epochs. By leveraging a pretrained language model here, we provide GDP with a strong understanding of medical language out of the box, so it can comprehend clinical narrative context and terminology.

**LLM Backbone and Fusion:** At the core of GDP is a transformer decoder architecture based on Meta's LLaMA model. We used a custom 3.2 billion-parameter LLaMA-like model as the backbone. For GDP-Non-Instruct, this backbone was initialized from a standard pre-trained LLaMA model (trained on general text but not instruction-tuned). For GDP-Instruct, we initialized from an instruction-tuned variant of LLaMA of the same size – essentially a model that had gone through an additional round of fine-tuning on instructional datasets (similar to how Alpaca or LLaMA-2 Chat models are derived). The backbone has the typical structure of a decoder-only transformer (layers of self-attention and feed-forward blocks), augmented with cross-attention to incorporate multimodal inputs. Specifically, we modified each transformer layer to include a cross-attention sublayer that attends to the sequence of structured data encoder outputs. In practice, the structured data encoder's final hidden states (the 50×256 sequence) are treated as "memory" for the LLM. At each generation step, the LLM's cross-attention heads can attend to any of those memory vectors, thereby allowing the content of the patient's timeline to influence the generated tokens. This mechanism is analogous to an encoder–decoder Transformer in machine translation, except here the "encoder" is the structured data module. Additionally, to fuse information from text notes when available (during fine-tuning for prediction), we employed a simple strategy: the 256-d note embedding from the text encoder is added to each time-step embedding of the structured sequence (broadcast across all positions) before it is fed into the cross-attention. This effectively informs the model that, for example, "this patient's sequence corresponds to someone whose note embedding is X," allowing the LLM to modulate its attention based on both structured and unstructured context. We also experimented with concatenating the note embedding as an additional token in the structured sequence, which yielded similar results. A layer normalization is applied to the fused sequence. The cross-attention layers in the LLM use multi-head attention to query this fused sequence from each decoder token position. In summary, during the generative mode, the LLM backbone takes as input the structured (and any text) context and generates a sequence of text tokens (the BHC) autoregressively, with cross-attention providing grounding. During the predictive mode, the LLM can be thought of as processing the structured (and text) data into an integrated latent representation (although in fine-tuning we do not actually *generate* text, we instead use the LLM's latent state for classification as described next).

**Multi-Task Output Heads:** For the predictive tasks, we added a lightweight classification head on top of the LLM backbone. After feeding the patient's data through the structured/text encoders and into the LLM (with no text prompt given to generate), we obtain the LLM's final hidden state for the sequence. We take the hidden state corresponding to a special beginning-of-sequence token (analogous to a [CLS] token) as an aggregate representation of the patient. This 256-d vector is then fed into a set of parallel linear classifier layers -one for each prediction task. Each classifier outputs a logit (for binary tasks) which is passed through a sigmoid to get a predicted probability. In our case, we had three classifiers (readmission, HF, T2DM ). The multi-task head thus shares the backbone representation but produces task-specific outputs. This design allows the model to leverage commonalities among tasks (e.g. features predictive of T2DM might overlap with those predictive of readmission) while still giving flexibility for each task's decision boundary. During fine-tuning, the model computes all task losses and sums them (with weighting) to update the shared layers. For the generative task, the output head is simply the LLM's native causal language modeling (LM) head – a linear layer tying into the LLM's vocabulary, used to predict the next token at each step of text generation.

Overall, GDP's architecture enables two training modes: a generative mode (structured data in, text out) with cross-attention guiding the text generation, and a classification mode (structured data in, label out) using the fused representation. The instruction-tuned vs non-tuned backbone differs only in initialization (and subsequent fine-tuning dynamics). By loading weights from a pre-trained LLaMA or LLaMA-Instruct model, we gave GDP a strong language foundation at the start, which we expected to help with textual coherence and general knowledge. The cross-attention layers ensured that this language model doesn't operate blindly, but rather always in context of the patient data – crucial for factual accuracy in generation.

## Model training

Generative pretraining stage: In the first stage, we trained the GDP model to model the joint distribution of EHR data and clinical text. Each training sample consisted of a patient's structured sequence (and note embedding, if note present) as context input, and the BHC text as the target output. The model was optimized to generate the BHC text from the context. We used a standard autoregressive language modeling loss (cross-entropy loss on each generated token) for this[40]. In addition, we simultaneously trained two auxiliary objectives on the structured data encoder: Masked Feature Prediction (MFP) and Next Time-step Prediction (NTP). For MFP, we randomly masked out a small fraction of the structured input vectors (or certain features within those vectors) and asked the model to reconstruct them. Specifically, we zeroed out some event embeddings and had a small decoder network predict the original values. The MFP loss was the mean squared error between the reconstructed and true feature values (or cross-entropy for masked code predictions). For NTP, for each sequence we withheld the final event vector and tasked the model to predict it given all prior events.

$$L_{MFP} = \frac{1}{|\mathcal{M}|} \sum_{i \in \mathcal{M}} ||x_i - \hat{x}_i||^2$$

$$L_{NTP} = ||h_T - x_{T+1}||^2$$

where,

$|\mathcal{M}|$ is the set of masked positions in the structured sequence.

$x_i$ is the original event embedding, $\hat{x}_i$ its reconstruction.

$h_T$ is the encoder's final hidden state for the first T events, and $x_{T+1}$ the true next-event embedding.

This was implemented via another feed-forward layer that takes the encoder's final hidden state (after CNN+Transformer) and outputs a prediction of the next event embedding; the NTP loss was the MSE between this prediction and the actual next-event vector.

We weighted these losses with coefficients $\lambda_{MFP}$ and $\lambda_{NTP}$ in the total pretraining loss. The overall pretraining objective was:

$$L_{Pretrain} = L_{CLM} + \lambda_{MFP} L_{MFP} + \lambda_{NTP} L_{NTP}$$

We set $\lambda_{MFP} = \lambda_{NTP} = 1$ initially (treating all losses equally), and we annealed them slightly down later in training to focus on LM loss as it improved. We used the AdamW optimizer with a learning rate of ~2e-4 for the backbone and 1e-3 for scratch parameters, and a linear warmup of a few thousand steps. Training was done on A100 GPU(80 GB) in mixed precision (fp16), with gradient accumulation such that the effective batch size was about 8 patient samples per update (each containing one note to generate). We trained until validation perplexity stopped improving (around 5 epochs through the data). The generative pretraining took around 2 days. Both GDP-Instruct and GDP-Non-Instruct underwent this pretraining (with identical hyperparameters), resulting in two pretrained model checkpoints.

Multi-task fine-tuning stage: In the second stage, we fine-tuned each pretrained model on the supervised prediction tasks. We froze the LLM backbone weights initially to avoid catastrophic forgetting of the language modeling capability. In the first 1–2 epochs of fine-tuning, we only trained the newly initialized components: the multi-task classification head and the fusion projections. We also unfroze the structured data encoder to adapt it to the new task signals. Then, we gradually unfroze the LLM backbone layers: over the next few epochs, we incrementally allowed more transformer layers to be trainable (starting from the top layers). This layer-freezing schedule, combined with a lower learning rate for the LLM layers (we used a 5× lower LR for backbone compared to encoders and heads), helped stabilize training. We fine-tuned using the AdamW optimizer (LR ~5e-5 for head, ~1e-5 for backbone). The loss function for fine-tuning was the sum of each task's binary cross-entropy loss. To handle class imbalance, we applied class-weighted focal loss for some tasks[41]. In practice, for readmission (where positives are relatively rare), using focal loss (γ=2) improved recall. The total fine-tune loss was

$$L_{finetune} = \sum W_i \, L_{task\ i}$$

where $w_k$ are task weights (we set them to 1 for each task). We trained in a multi-task fashion: each batch contained examples with all task labels (if a label was missing for a patient, we masked that loss). Fine-tuning continued for ~10 epochs, with early stopping based on validation AUROC to prevent overfitting. The final chosen model was the epoch with highest average AUROC across tasks on the validation set. We repeated fine-tuning for both GDP-Instruct and GDP-Non-Instruct starting from their respective pretrained weights.

For the summarization task (BHC generation), we fine-tuned GDP-Instruct and GDP-Non-Instruct separately on note generation as well (though they had already seen that task in pretraining). We found that a brief additional fine-tuning on the smaller supervised set of paired structured data–BHC (using only the LM loss) helped alignment with the reference style and improved ROUGE slightly. The baselines (Meta Llama and Med42) were also fine-tuned on this task using the same hyperparameters for a fair comparison.

## Evaluation metrics and analysis

We evaluated classification performance on the test set using several metrics. Discrimination was measured by the area under the ROC curve (AUROC) and area under the precision–recall curve (AUPRC), computed for each task. These threshold-independent metrics summarize the model's ability to rank-order predictions. We calculated 95% confidence intervals for AUROC/AUPRC via bootstrapping 1000 samples on test set. Threshold-dependent metrics were computed at the conventional 0.5 probability cutoff (since we had balanced validation sets, this was reasonable). We report F1score, precision, recall, and accuracy for each model on each task (Table 1). F1 is the harmonic mean of precision and recall, reflecting a balance between sensitivity and specificity.

$$Precision = \frac{TP}{TP + FP}$$

$$Recall = \frac{TP}{TP + FN}$$

$$F_1 = 2 \times \frac{\text{Precision} \times \text{Recall}}{\text{Precision} + \text{Recall}}$$

$$Accuracy = \frac{TP + TN}{TP + TN + FP + FN}$$

$$AUROC = \int_0^1 \text{TPR}\left(\text{FPR}^{-1}(t)\right) dt$$

$$AUPRC = \int_0^1 \text{Precision}\left(\text{Recall}^{-1}(r)\right) dr$$

Where TP / TN/ FP / FN denote true positives, true negatives, false positives, and false negatives, respectively. TPR = TP/(TP + FN) and FPR = FP/(FP + TN). The integrals for AUROC and AUPRC are typically estimated via trapezoidal approximation over the ROC and precision–recall curves.

For generative performance, we used standard natural language generation (NLG) metrics on the test set of discharge summary. We computed ROUGE-1, ROUGE-2, and ROUGE-L F1-scores (covering unigram overlap, bigram overlap, and longest common subsequence overlap) between the generated summaries and reference summaries. We report the F-measure variant of ROUGE, which balances precision and recall of overlap. We also calculated BLEU scores (up to BLEU-4, cumulative) which measure n-gram precision – an indicator of how exactly the model reproduced the reference phrasing. However, since strict n-gram matches can be overly harsh for this task (there can be many ways to write the same clinical fact), we included BERTScore, which uses a pre-trained language model to assess semantic similarity between the generated and reference text. BERTScore outputs a similarity score in [0,1]. All these metrics were computed using standard libraries (rouge-score, nltk for BLEU, and the BERTScore package with the recommended biomedical BERT model). We considered the reference summaries written by clinicians as the gold standard, and we averaged the scores across all test samples for each model (Table 2). To test for significance in metric differences, we used paired bootstrap resampling (e.g., for ROUGE-L differences between GDP-Instruct and baseline LLM).

$$ROUGE\_N\_F1 = 2 \times \frac{Overlap\_Precision\_N \times Overlap\_Recall\_N}{Overlap\_Precision\_N + Overlap\_Recall\_N}$$

$$BLEU = BP \times \exp\left(\sum_{n=1}^{4} w_n \ln p_n\right)$$

$$BERTScore\_F1 = 2 \times \frac{Precision\_BERT \times Recall\_BERT}{Precision\_BERT + Recall\_BERT}$$

Where,

$$\text{Overlap\_Precision}_N = \frac{\sum \text{matched N-grams}}{\sum \text{generated N-grams}}$$

$$\text{Overlap\_Recall}_N = \frac{\sum \text{matched N-grams}}{\sum \text{reference N-grams}}$$

$$p_n = \text{Overlap\_Precision}_n$$

$$\text{Precision}_{\text{BERT}} = \frac{1}{L}\sum_{i=1}^{L} \max_{j} \text{sim}(t_i, s_j)$$

$$\text{Recall}_{\text{BERT}} = \frac{1}{M}\sum_{j=1}^{M} \max_{i} \text{sim}(s_j, t_i)$$

$$BP = \begin{cases} 1, & \text{if } c > r \\ \exp(1 - r/c), & \text{if } c \leq r \end{cases}$$

For human evaluation, as described in Results, two experts rated a set of 200 generated summaries from each model. The summaries were randomly selected such that each evaluator saw each model's output for the *same* patient (allowing within-patient comparison across models, without knowing which was which). The rating criteria (faithfulness, completeness, conciseness, readability, and clinical utility) were each scored 1–5. We also computed Cohen's κ to measure inter-rater reliability for each criterion; all κ were in the 0.6–0.75 range, indicating substantial agreement. No serious disagreements (difference >2 on the Likert scale) occurred on the sampled evaluations.

Finally, for the interactive chat evaluation, we did not perform a formal study but rather anecdotal trials. We did ensure that the chat interface had access to the model's weights and that the patient data was loaded in a consistent way for both GDP-Instruct and GDP-Non-Instruct. We logged example Q&A sessions for qualitative analysis. These examples were used to subjectively assess the coherence and accuracy of each model's responses when faced with questions about the patient's data. While not a quantitative test, it provided insights reported in the Results to illustrate the models' capabilities in a conversational context.

All experiments were conducted in Python using PyTorch. Model training and evaluation code will be released to allow reproduction of these results.

## Author contributions

S.S. conceptualized, designed, and organized this study, analyzed the results, and wrote, reviewed, and revised the paper. H.Z., Y.J., M.B., X.W., C.L., M.G.K., S.V. analyzed the results, and wrote, reviewed, and revised the paper. Y.W. conceptualized, designed, and directed this study, wrote, reviewed, and revised the paper.

## Funding

The research reported in this article was partially supported by the National Institutes of Health awards UL1 TR001857, U24 TR004111, and R01 LM014306. The sponsors had no role in study design, data collection, analysis, interpretation, report writing, or decision to submit the paper for publication.

## Ethics statement

Use of the MIMIC-IV database was approved by the institutional review boards of the Massachusetts Institute of Technology and Beth Israel Deaconess Medical Center, with a waiver of informed consent because the dataset is de-identified. All methods were carried out in accordance with relevant guidelines and regulations.

## Data availability

The MIMIC-IV dataset used in this study is available through PhysioNet with credentialed access. Processed data in the MEDS JSONL format and task labels can be made available from the corresponding authors upon reasonable request (and in compliance with data use agreements). The discharge summary texts are part of MIMIC-IV and subject to the same access requirements.

## Code availability

The code for the GDP model architecture, training procedures, and evaluation scripts will be released on GitHub (link to be provided upon publication). This repository will include instructions to reproduce the experiments and will reference the specific versions of libraries and pretrained models used.

## Competing interests

The authors declare no competing interests.

## Acknowledgments

We thank our evaluators who meticulously did human evaluation.

# Supplementary Material to

# Generative Foundation Model for Structured and Unstructured Electronic Health Records

**Sonish Sivarajkumar, BS[1], Hang Zhang, MS[1], Yuelyu Ji, MS[1]; Maneesh Bilalpur, MS[1]; Xizhi Wu, MS[2]; Chenyu Li, MS[3,]; Min Gu Kwak, PhD[2]; Shyam Visweswaran, MD, PhD[1,3,4], Yanshan Wang, PhD[1,2,3,4,5] †**

[1]Intelligent Systems Program, University of Pittsburgh, Pittsburgh, PA, USA
[2]Department of Health Information Management, University of Pittsburgh, Pittsburgh, PA, USA
[3]Department of Biomedical Informatics, University of Pittsburgh, Pittsburgh, PA, USA
[4]Clinical and Translational Science Institute, University of Pittsburgh, Pittsburgh, PA
[5]Hillman Cancer Center, University of Pittsburgh Medical Center, Pittsburgh, PA
**† corresponding author: Yanshan Wang, yanshan.wang@pitt.edu**

## 1. Introduction

This Supplementary Material provides comprehensive details that were omitted from the main manuscript due to space constraints. It is organized into three parts: (1) Expanded Model Architecture (2) Key hyperparameters and (3) Qualitative Evaluation Results, which is an extension to the results section from the main manuscript.

## 2. Model Architecture

### 2.1 Structured Data Encoder

The structured data encoder processes patient timeline data using a combination of CNN and Transformer architecture. Each patient timeline consists of 100 hourly embeddings with 128 dimensions per embedding.

The CNN component consists of three sequential convolutions, each with a kernel size of 3, a stride of 1, and a padding of 1. The first convolutional layer maintains 128 channels and is followed by batch normalization and ReLU activation. The second layer expands to 256 channels, also with batch normalization and ReLU activation. The third layer maintains 256 channels but employs layer normalization followed by GELU activation. The CNN produces a sequence representation of dimensions $100 \times 256$.

The Transformer component then processes the CNN output. Learned positional embeddings of matching dimensions are added to the sequence representation before passing it through four Transformer encoder layers. Each Transformer layer utilizes a hidden dimension of 256, 8 attention heads, a feed-forward network with an inner dimension of 512, and a dropout rate of 0.1, with pre-normalization architecture. The final output maintains the dimensions of $100 \times 256$. A 256-dimensional demographic embedding is added to each time step before fusion with other modalities, ensuring incorporation of patient-specific static features.

Each patient timeline consists of 100 hourly embeddings (128 D). A 1D-CNN front end applies three convolutions (kernel = 3, stride = 1, padding = 1):

- Conv1: 128→128 channels, BatchNorm → ReLU
- Conv2: 128→256 channels, BatchNorm → ReLU
- Conv3: 256→256 channels, LayerNorm → GELU

This produces a sequence of shape [100 × 256]. Learned positional embeddings (100 × 256) are added, and the result is passed through four Transformer encoder layers (hidden = 256, 8 heads, FFN = 512, dropout = 0.1, pre-norm). The final output is [100 × 256]. A 256-D demographic embedding is added to each time step before fusion.

## 2.2 Text Encoder

The text encoder processes clinical notes using BioClinicalBERT, which was pre-trained on MIMIC-III data. Clinical notes up to 512 tokens are processed through this model, which is based on the bert-base-uncased architecture. The [CLS] token embedding (768 dimensions) is extracted as a representation of the entire note and linearly projected to 256 dimensions without activation. This representation is then broadcast and added to each of the 100 time steps from the structured data encoder.

Clinical notes (≤512 tokens) are encoded with BioClinicalBERT (bert-base-uncased pretrained on MIMIC-III). The [CLS] embedding (768 D) is linearly projected to 256 D (no activation). This 256-D vector is broadcast and added to each of the 100 structured embeddings alongside demographics.

## 2.3 LLaMA Decoder Backbone and Cross-Modal Fusion

The LLaMA decoder backbone serves as the core of the GDP model, with two variants implemented: GDP-Non-Instruct utilizing a raw 3.2B parameter LLaMA checkpoint, and GDP-Instruct using an instruction-tuned variant of the same size.

Each decoder consists of 24 layers incorporating masked self-attention, cross-modal attention, and feed-forward components. The masked self-attention mechanism employs a hidden dimension of 256, 16 attention heads, causal masking, and a dropout rate of 0.1. The cross-attention mechanism processes queries from the self-attention 256-dimensional output against keys and values derived from the fused embeddings ($100 \times 256$ dimensions), utilizing 8 attention heads and a dropout rate of 0.1. The feed-forward network expands the representation from 256 to 1024 dimensions before projecting back to 256, employing GELU activation and a dropout rate of 0.1.

LayerNorm is applied before each sublayer, with residual connections employed throughout the network. The vocabulary remains the original LLaMA 32,000-token BPE. During autoregressive generation, the decoder attends to the fused EHR embeddings at every generation step. For classification tasks, the hidden state corresponding to the beginning-of-sequence token is utilized.

We use two LLaMA-3.2B variants:

- GDP-Non-Instruct (raw 3.2 B checkpoint)
- GDP-Instruct (instruction-tuned variant)

Each of 24 decoder layers comprises:

1. Masked self-attention (hidden = 256, 16 heads, causal mask, dropout = 0.1)
2. Cross-attention (queries = 256-D from self-attention; keys/values = [100 × 256] fused embeddings; 8 heads, dropout = 0.1)
3. Feed-forward (256 → 1 024 → 256, GELU, dropout = 0.1)

Preceding each sublayer is LayerNorm, and residual connections are employed. Vocabulary remains LLaMA's 32 000-token BPE. During autoregressive generation, the decoder attends to fused EHR embeddings at every step. For classification, the hidden state at the <BOS> token is used.

## 2.4 Auxiliary Pretraining Objectives

To encourage the model to learn both structured-to-text alignment and internal representation of EHR patterns, we employ three objectives during generative pretraining:

1. **Language Modeling (LM) on BHC**
   We minimize the standard autoregressive cross-entropy:

   $$L_{LM} = -\sum_{t=1}^{L} \log P_\theta(w_t \mid w_{<t}, \text{EHR}_{\text{fused}})$$

   where $w_t$ are tokens of the Brief Hospital Course, and Here, $L \leq 256$ and $EHR_{fused}$ denotes the 100×256 cross-attention memory.

2. **Masked Feature Prediction (MFP)**
   For 15 % of the 100 time steps, we apply masking per the following scheme:

   - 80 % probability: Zero out the entire 128-dim event embedding.
   - 10 % probability: Replace only categorical code sub-embeddings with a learned [MASK] vector (keeping numeric buckets intact).
   - 10 % probability: Leave the embedding unchanged (to avoid overfitting to the mask token).
   
   A two-layer MLP (256 → 256 → 128) then attempts to reconstruct the original 128-dim event embedding at each masked position iii. We compute:

   $$L_{MFP} = \frac{1}{|M|} \sum_{i \in M} \|\hat{s}_i - s_i^{\text{true}}\|^2$$

   where M is the set of masked time-step indices.

3. **Next Time-Step Prediction (NTP)**
   We withhold the final time-step embedding $s_{100}$ from the input. After encoding $s_{1:99}$ through the CNN→Transformer, we take the last hidden state $h_{99} \in R^{256}$ and pass it through a linear layer (256 → 128) to predict $\hat{s_{100}}$. The loss is

   $$L_{NTP} = \|\hat{s_{100}} - s_{100}^{\text{true}}\|^2$$

The overall pretraining objective is:

$$L_{\text{pretrain}} = L_{LM} + \lambda_{MFP} L_{MFP} + \lambda_{NTP} L_{NTP}$$

where $\lambda_{MFP} = \lambda_{NTP} = 1.0$. Initially, decaying linearly to 0.5 after epoch 3.

## 2.5 Training Configurations

The GDP model is trained using a two-staged approach: generative pretraining followed by supervised fine-tuning. For all training stages, we use the AdamW optimizer with $\beta_1 = 0.9$, $\beta_2 = 0.95$, $\epsilon = 10^{-8}$ and weight decay of 0.1. The learning rate follows a cosine schedule with linear warmup.

During generative pretraining, we use an effective batch size of 8 through gradient accumulation of four micro-batches of size 2. The backbone learning rate is set to $2 \times 10^{-4}$ while encoders and prediction heads use a higher rate of $1 \times 10^{-3}$. The learning rate warms up linearly over the first 1,000 steps and then decays following a cosine schedule to zero over 50,000 steps (approximately 5 epochs). This stage is conducted using mixed precision on A100 GPUs with early stopping if validation perplexity on BHC fails to improve for two consecutive epochs.

For multi-task fine-tuning, we simultaneously optimize for three clinical prediction tasks: Heart Failure (HF), Type 2 Diabetes (T2DM), and 30-day Readmission. The beginning-of-sequence (BOS) hidden state with 256 dimensions feeds three parallel classification heads, each consisting of a linear layer followed by sigmoid activation. We use a batch size of 16 with learning rates of $5 \times 10^{-5}$ for fusion components and classification heads, and $1 \times 10^{-5}$ for the backbone. The fine-tuning loss is defined as:

(insert equation)

where BCE refers to Binary Cross-Entropy loss, $w_{HF} = 0.90$, $w_{DM} = 0.86$, and $w_{Readm} = 0.85$ are the balancing hyperparameters. Focal Loss hyperparameters are set to $\gamma = 2$ and $\alpha = 0.25$.

To prevent overfitting during fine-tuning, we implement a progressive unfreezing strategy. During the first two epochs, the decoder remains frozen. For epochs three through five, we unfreeze the top six layers. Finally, all parameters are trainable for the remaining epochs. We use early stopping based on mean validation AUROC with a patience of three epochs and a maximum of ten epochs.

**Generative Pretraining:**

Optimizer: AdamW (backbone LR = $2\times10^{-4}$; encoders + heads LR = $1\times10^{-3}$; weight decay = 0.01; $\beta_1$ = 0.9, $\beta_2$ = 0.999, $\varepsilon$ = $10^{-8}$). Warmup: linear over 1 000 steps; cosine decay to zero over 50 000 steps (~5 epochs). Batch size: effective = 8 (gradient accumulation of four micro-batches of 2). Mixed precision (FP16) on A100 GPUs. Early stopping if validation perplexity on BHC fails to improve for two consecutive epochs.

**Multi-Task Fine-Tuning:**

Tasks: HF, T2DM , and 30-day Readmission. The <BOS> hidden state (256 D) feeds three parallel linear → sigmoid heads. Loss:

$$L_{fine} = w_{HF}\,BCE(p_{HF}, y_{HF}) + w_{T2DM}\,BCE(p_{T2DM}, y_{T2DM}) + w_{Readm}\,Focal(p_{Readm}, y_{Readm})$$

Optimizer: AdamW (fusion + heads LR = $5\times10^{-5}$; backbone LR = $1\times10^{-5}$; weight decay = 0.01). Freezing schedule: epochs 1–2 freeze decoder, epochs 3–5 unfreeze top 6 layers, epochs 6–10 unfreeze all. Batch size = 16. Early stopping on mean validation AUROC (patience = 3, max epochs = 10).

## 3. Key Hyperparameters

Below is a concise table summarizing the principal hyperparameters used in both pretraining and fine-tuning. For full details (including less critical settings such as tokenizer specifics and data split random seeds), refer to the code repository and configuration files.

**Table S1**. Key hyperparameters

| Component | Hyperparameter | Value / Setting |
|---|---|---|
| Structured Encoder (CNN → Trans.) | CNN Layer 1 filters (kernel = 3) | 128 channels → ReLU |
| | CNN Layer 2 filters (kernel = 3) | 256 channels → ReLU |
| | CNN Layer 3 filters (kernel = 3) | 256 channels → GELU |
| | Transformer layers | 4 |
| | Transformer hidden dimension | 256 |
| | Transformer heads | 8 |

|  | Transformer FFN inner dimension | 512 |
|  | Transformer dropout | 0.1 |
| Text Encoder (BioClinicalBERT) | Pretrained checkpoint | bioclinicalbert-base-cased |
|  | Projection to fusion dimension | 768 → 256 |
| LLM Decoder Backbone | Model variant | LLaMA-3.2B (Non-Instruct / Instruct) |
|  | Number of decoder layers | 24 |
|  | Decoder hidden dimension | 256 |
|  | Decoder attention heads | 16 |
|  | Decoder feed-forward dimension | 1,024 |
|  | Decoder dropout | 0.1 |
| Auxiliary Objectives | Masking ratio (MFP) | 15 % |
|  | $\lambda_{MFP}$ | 1.0 (initial), 0.5 (after epoch 3) |
|  | $\lambda_{NTP}$ | 1.0 (initial), 0.5 (after epoch 3) |
| Pretraining Optimizer | Algorithm | AdamW |
|  | LR (backbone) | $2\times10^{-4}$ |
|  | LR (encoders + new heads) | $1\times10^{-3}$ |
|  | Weight decay | 0.01 |
|  | Warmup steps | 1,000 |
|  | Total steps | ≈ 50,000 (~ 5 epochs) |
|  | Effective batch size | 8 (admissions × 4 micro-batches) |
|  | Precision | FP16 (mixed-precision) |
| Fine-Tuning Optimizer | Algorithm | AdamW |
|  | LR (fusion + classifier heads) | $5\times10^{-5}$ |
|  | LR (backbone) | $1\times10^{-5}$ |
|  | Weight decay | 0.01 |
|  | Freezing schedule | Epochs 1–2: freeze decoder; Epochs 3–5: unfreeze top 6; Epochs 6–10: all trainable |
|  | Batch size | 16 (admissions) |
|  | Early stopping (patience) | 3 epochs (mean validation AUROC) |
| Multi-Task Loss Weights | $w_{HF}$ | 0.90 (positive rate ≈ 11 %) |

|  | w_T2DM | 0.86 (positive rate ≈ 14 %) |
|--|--------|------------------------------|
|  | w_Readm | 0.85 (positive rate ≈ 15 %) |
|  | Readmission loss | Focal (γ = 2, α = 0.25) |

## 4. Data types and encoding

**Table S2.** Structured data types and encoding approaches

| Data type | Encoding / handling |
|-----------|---------------------|
| Age, sex (demographics) | Directly encoded (age normalized, sex as binary categorical) |
| Diagnoses (ICD-9/10) | Mapped to integer indices; embedded |
| Procedures | Mapped to integer indices; embedded |
| Lab results (numeric) | Z-scored or bucketed into ranges |
| Lab results (categorical, e.g., positive/negative) | One-hot encoded / mapped to integer |
| Medications – oral/tablets (scheduled) | Encoded as discrete events with time gaps |
| Medications – IV/continuous | Represented as time-stamped events with duration |
| Vital signs (BP, HR, RR, etc.) | Continuous values; forward-filled within windows |
| Input/output measures | Continuous values; normalized |
| Ventilator settings and other device parameters | Continuous/categorical as appropriate |

## 5. Ablation analysis: impact of instruction tuning and auxiliary objectives

The divergent results between GDP-Instruct(GDP with LLaMA-instruct backbone) and GDP-Non-Instruct(GDP with LLaMA-non-instruct backbone) highlight an interesting trade-off. The standard LLM backbone (Non-Instruct) proved slightly superior for purely predictive tasks, whereas the instruction-tuned backbone was markedly superior for text generation and summary tasks. This suggests that the initial training paradigm of the LLM influences what it excels at: instruction tuning endows the model with strengths in following prompts and generating fluent, contextually appropriate text (critical for narrative tasks), but the process might marginally dilute the predictive signal for classification tasks. *(For clarity: fine-tuning refers to updating a pre-trained model's parameters on a specific task or domain; instruction tuning is a type of fine-tuning where the model learns to follow human instructions across diverse commands; and prompt engineering means crafting effective inputs for an already-trained model without changing its parameters).* In our experiments,

GDP-Non-Instruct consistently achieved higher AUROC/AUPRC on structured tasks (Table 1), indicating a better discrimination capacity after fine-tuning, while GDP-Instruct consistently produced more accurate and readable free-text outputs (Tables 2 and 3). In practice, both variants still performed strongly on all tasks, but the choice of LLM backbone could be optimized based on the target application (analytic vs. generative). Future work may explore hybrid approaches to get the best of both – for example, further instruction-tuning the model after fine-tuning on predictive tasks, or using lightweight adapters to re-balance these capacities.

We also investigated the contribution of GDP's auxiliary pretraining losses – the Masked Feature Prediction (MFP), inspired by Masked Language Modeling [42], and Next Time-step Prediction (NTP), inspired by Next Token Prediction [43], tasks – to its performance(Table 4). These losses were applied during generative pretraining to encourage the structured data encoder to learn richer representations. Qualitatively, we found they were vital for learning temporal relationships. Without NTP, for instance, the model's ability to anticipate disease progression events was weakened, leading to drops in predictive accuracy. An ablation experiment where we removed the NTP loss during pretraining resulted in a noticeable decrease in AUROC (e.g., ~3–5 point drop on HF and T2DM prediction) and a flatter precision–recall curve, indicating the model was less adept at identifying patients with those conditions. MFP had a more modest but still positive effect: it helped the model capture fine-grained details in the EHR vectors, which translated to slight improvements in recall for the diagnosis tasks (by ensuring the encoder doesn't ignore small-but-important code features).

**Table S3.** Ablation of Auxiliary Objectives on Discriminative Tasks

| Model Variant | AUROC (HF) | AUPRC (HF) | AUROC (T2DM) | AUPRC (T2DM) |
|---|---|---|---|---|
| Full (MFP + NTP) | 0.923 | 0.710 | 0.817 | 0.707 |
| – NTP only | 0.893 | 0.674 | 0.795 | 0.682 |
| – MFP only | 0.915 | 0.697 | 0.803 | 0.695 |
| – MFP & NTP | 0.870 | 0.640 | 0.780 | 0.660 |

Concurrently training on MFP and NTP, alongside the main language modeling objective, clearly forced the model to attend the structured data in a way that pure language modeling would not. This is evident from the baseline comparisons: models like CEHR-BERT and CLMBR also leverage sequential structure (CEHR-BERT via a Transformer over code sequences, CLMBR via next-code prediction). Yet GDP's joint training with a generative objective and explicit temporal self-supervision appears to instill a more integrated understanding. In essence, GDP's pretraining strategy – generate text conditioned on EHR while predicting masked and next events – produced a multifaceted representation that benefited both types of tasks. Our results support this: GDP-Instruct outperformed a similarly-sized instruct-tuned LLMs that were not pretrained on structured EHR data (Table 2 and 3 baselines), indicating that grounding text generation in patient timelines (through cross-attention and auxiliary losses) yields more relevant and factual narratives than a generic medical LLM fine-tuned on the same summaries. Meanwhile, GDP-Non-Instruct's substantial boost over baselines like FPM or CLMBR on predictive tasks demonstrates the value of combining language-model pretraining with temporal encoding. We note that CEHR-BERT and CLMBR were strong in their domains, but lacked a generative text pretraining component; GDP's unique fusion of these elements likely explains its superior overall performance.

## 6. Qualitative evaluation via interactive chat

Beyond static benchmarks, we explored GDP's potential in an interactive EHR question-answering setting. We developed a prototype chat interface where a user (e.g. a clinician, or health informatics specialist) can query the model about a patient's case after "loading" that patient's EHR data into the model's context. Using held-out test patient records, we observed how the two GDP variants responded to ad-hoc questions. For example, given a patient's structured data and notes indicating an admission for pneumonia, we asked: "Q: What were the main events during this hospital stay?" GDP-Instruct responded with a concise summary: "The patient was admitted with a high fever and cough. They were started on IV antibiotics for pneumonia, required supplemental oxygen briefly, and improved over 3 days. Key events include a fever spike to 39.2°C on day 1, blood cultures growing Streptococcus pneumoniae, and transition from IV to oral antibiotics on day 4 before discharge." In contrast, GDP-Non-Instruct produced a less organized answer: "Pneumonia. They got antibiotics. Fever went up (39.2). Blood culture positive for strep pneumo. O given." Both are factually correct, but the instruct model's answer was more fluent and provided a clearer narrative flow, likely reflecting its training to generate human-like summaries. We also asked follow-up questions such as "Q: What follow-up care is recommended?". GDP-Instruct was able to infer and respond with a reasonable recommendation (e.g., "Follow-up with the primary care physician in 1 week, repeat chest X-ray in 4-6 weeks, and continue oral antibiotics for the prescribed course."), whereas GDP-Non-Instruct tended to give more generic or hesitant answers (sometimes just restating to follow up with a doctor without specifics). These anecdotal tests underscore that instruction tuning imparts better conversational abilities – GDP-Instruct was adept at understanding the question and formulating a contextually appropriate answer using the EHR data, effectively acting like a clinical assistant. GDP-Non-Instruct, while containing the necessary information in its latent representation, was less able to articulate it without direct prompting. This qualitative use-case highlights a future application of models like GDP: interactive, EHR-aware decision support systems. It also shows the value of bridging generative and predictive capacity – the model not only predicts risks, but can explain and discuss a patient's case in natural language. We note, however, that these chat observations were not rigorously quantified and are meant to inspire further evaluation on clinical QA benchmarks, which is the future plan.